\documentclass[10pt,conference,a4paper]{IEEEtran}
\usepackage{cite}
\usepackage{amsmath,amssymb,amsfonts}
\usepackage[utf8]{inputenc}

\usepackage{siunitx}
\usepackage{algorithmic}
\usepackage{graphicx}
\usepackage{textcomp}
\usepackage{xcolor}

\def\BibTeX{{\rm B\kern-.05em{\sc i\kern-.025em b}\kern-.08em
    T\kern-.1667em\lower.7ex\hbox{E}\kern-.125emX}}
\begin{document}

\title{HP\textsuperscript{2}IFS: Head Pose estimation exploiting Partitioned Iterated Function Systems\\}

\author{\IEEEauthorblockN{Carmen Bisogni}
\IEEEauthorblockA{\textit{Dept. of Computer Science} \\
\textit{University of Salerno}\\
Fisciano, Italy \\
cbisogni@unisa.it}
\and
\IEEEauthorblockN{Michele Nappi}
\IEEEauthorblockA{\textit{Dept. of Computer Science} \\
\textit{University of Salerno}\\
Fisciano, Italy \\
mnappi@unisa.it}
\and
\IEEEauthorblockN{Chiara Pero}
\IEEEauthorblockA{\textit{Dept. of Computer Science} \\
\textit{University of Salerno}\\
Fisciano, Italy \\
cpero@unisa.it}
\and
\IEEEauthorblockN{Stefano Ricciardi}
\IEEEauthorblockA{\textit{Dept. of Biosciences and Territory} \\
\textit{University of Molise}\\
Pesche, Italy \\
stefano.ricciardi@unimol.it}
}

\maketitle

\begin{abstract}
Estimating the actual head orientation from 2D images, with regard to its three degrees of freedom, is a well known problem that is highly significant for a large number of applications involving head pose knowledge. Consequently, this topic has been tackled by a plethora of methods and algorithms the most part of which exploits neural networks. Machine learning methods, indeed, achieve accurate head rotation values yet require an adequate training stage and, to that aim, a relevant number of positive and negative examples. In this paper we take a different approach to this topic by using fractal coding theory and particularly Partitioned Iterated Function Systems to extract the fractal code from the input head image and to compare this representation to the fractal code of a reference model through Hamming distance. According to experiments conducted on both the BIWI and the AFLW2000 databases, the proposed PIFS based head pose estimation method provides accurate yaw/pitch/roll angular values, with a performance approaching that of state of the art of machine-learning  based algorithms and exceeding most of non-training based approaches.\newline
\end{abstract}


\section{Introduction}
Whenever an image containing a human face is captured and analyzed, determining head pose often represents a crucial and challenging task for a wide number of applications such as biometric recognition or video surveillance, just to name the most obvious. This particular kind of intra-class variation can significantly affect recognition accuracy as well as have an impact  on any application requiring the normalization of head rotation with respect to its three degrees of freedom. The generalization of this problem from single image analysis to video sequence analysis, further extend its relevance and reach, which is typically proportional to absolute head rotation values. 
Not surprisingly, then, head pose estimation represents a  hot research topic that has been approached by a large number of methods, most of which are based on the use of convolutional neural networks (CNNs) and deep learning (DL) techniques, often providing high accuracy with the not negligible cost of a prior training stage requiring positive and negative examples as input stimula.
Aiming at rivalling the accuracy of CNNs and DL based approaches, yet without the need of previous training process, we present HP2IFS a novel Head Pose estimation method based on Partitioned Iterated Function Systems (PIFS) and fractal image coding. PIFS, typically used for "lossy" image compression applications, is exploited as a mean to encode auto-similarities within the face image. The resulting fractal parameters vector is therefore compared to an angular reference array previously built through the same fractal coding algorithm by means of Hamming distance metric. Experiments conducted on both the Kinect Head Pose Database (BIWI) and the Automated Facial Landmarks in the Wild dataset (AFLW2000), confirm the validity of the proposed approach and its high accuracy compared to the state-of-the-art. The rest of this paper is organized as follows. Section II. resumes relevant works related to the present method. Section III. recalls the basics of fractal image coding and PIFS, while Section IV. describes in detail the proposed approach. Finally, Section V. presents the results of the experiments conducted and Section VI. concludes drawing some directions for future research.

\begin{figure}[!ht]
\centering
  \includegraphics[width=.48\textwidth]{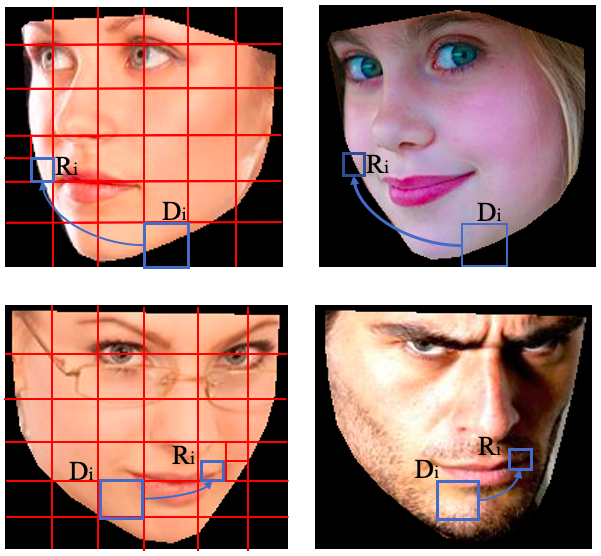}
  \caption{The transformations between the domain blocks (D) and the range blocks (R), on image pairs featuring similar angular values.}
  \label{fig:codec1}
\end{figure}

\section{Related Works}

As already outlined in the introduction, a plethora of head pose estimation methods has been proposed throughout the last decades according to literature surveys such as \cite{trivedi2009} and \cite{Czup2014}. The majority of contributions to this topic have been based on machine learning approaches through deep neural architectures and more eminently on CNNs which are particularly suited to image processing. First works exploited Gaussian processes \cite{rasmussen2004gaussian}, or partial least square (PLS) regression techniques \cite{abdi2003partial} possibly implemented on support vector machines (SVMs) \cite{smola2004tutorial}. More recently, Drouard et al. \cite{drouard2017robust} proposed an approach to regression aimed at robust head pose estimation in the presence of intra-class variations, by mapping feature vectors associated to bounding boxes of faces onto the space of pose-angles and boundig-box translations. In \cite{kumar2017kepler} face keypoints detection and pose prediction are achieved by learning efficient regressors through an original neural architecture named Heatmap CNN, capable of extracting both global and local features to improve prediction reliability. Aiming at simultaneous face detection, keypoints localization, pose estimation and gender recognition, Ranjan et al. \cite{ranjan2019hyperface} make full use of deep CNN to develop a synergetic approach to these tasks that result in a boost up of each single task performance.
A total 3D solution to face alignment even for large poses (greater than 45° and up to 90°) is proposed by Zhu et al. \cite{zhu2017face} fitting 3D Morphable Models to 2D input face image by means of a cascaded CNN.
With regard to assessing the limits of deep NN for face alignment problem on existing 2D and 3D public datasets, the authors \cite{bulat2017far} develop a new public 3D facial landmark dataset named LS3D-W unifying all existing datasets and study the performance of both 2D and 3D pose estimation networks in this new context.
In the quest for a method not relying on facial keypoints, Ruiz et al. \cite{finegrained} propose to train a multi-loss CNN on the 300W-LP sinthetic dataset to achieve yaw/pitch/roll prediction through joint binned pose classification/regression from intensity images.
On a similar line of research, the authors of \cite{yang2019fsa} introduce a method without landmarks named FSA-Net, which relies on a soft stagewise regression scheme combining feature maps from different layers/stages.
In \cite{hsu2018quatnet}, a method combining L2 regression loss combined with an ordinal regression loss is used to train a CNN for highly accurate face pose estimation by addressing the ambiguity related to Euler angle representation through quaternions.
Wang et al. \cite{wang2019deep} exploits DL in a coarse-to-fine strategy based on two sub-networks trained jointly, where the coarse stage classifies the input face into four bins and the fine regression stage delivers pose angles.
Among the few works not based on machine learning methods, in \cite{Ricciardi_pose} a quad-tree based vector representing facial landmarks locations is compared to a pre-computed set of 2223 angular variations to estimate head pose to determine the closest approximation to the actual head pose.

The proposed HP\textsuperscript{2}IFS method, besides not requiring any prior training, not even requires any landmark detection, since it only relies on intrinsic image auto-similarities.

\section{Fractal Image Coding Basics}
\label{sec_frac}
The fundamental principle of fractal coding consists of the representation of an image by a contractive transform of which the fixed point is close to that image. This fixed point is named as \textit{fractal} \cite{fisher1995}.  Banach’s fixed point theorem guarantees that, within a complete metric space, the fixed point of such a transform may be recovered by iterated application thereof to an arbitrary initial element of that space \cite{distasi2005range}.
Fractal coding techniques are based on the collage theorem, which  provides a bound on the distance between the image to be encoded and the fixed point of a transform, in terms of the distance between the transform of the image and the image itself. In the early 1980s, Barnsley \cite{barnsley1993fractal} applied his knowledge of fractal and mathematics to image compression, explaining that this transform was composed of the union of a number of affine mappings on the entire image - an iterated function system (\textit{IFS}) \cite{fisher1995}. 
Fractal compression became a practical reality with the introduction by Jacquin of the \textit{partitioned}-IFS (\textit{PIFS}), which differs from an IFS in that each of the individual mappings operates on a subset of the image, rather than on the entire image \cite{fisher1995}. 

\noindent The basic principle is that an image can be reconstructed by using the self-similarities in the image itself. When encoding an image, the algorithm partitions the original image into non-overlapping domain regions; after this, a new partition into smaller blocks (range blocks) is applied. For every range block, the best matching domain block is searched among all domain blocks by performing a set of transformations on the blocks. The compression is obtained by storing only the descriptions of these transformations.

\noindent A number of algorithms have been proposed by various authors to increase the efficiency of this method; the differences between the majority of existing fractal coding schemes may be classified into the following categories \cite{distasi2005range}:
\begin{itemize}
    \item partition imposed on the image support by the range blocks;
\end{itemize}
\begin{itemize}
    \item composition of the pool of domain blocks;
\end{itemize}
\begin{itemize}
    \item class of transforms applied to the domain blocks;
\end{itemize}
\begin{itemize}
    \item type of search used in locating suitable domain blocks;
\end{itemize}
\begin{itemize}
    \item representation and quantization of the transform parameters.
\end{itemize}

\subsection{Fractal Image Coding Implementation}
\noindent The fractal image encoding algorithm \cite{2015hybrid} works, in principle, as follows:\\

\noindent \textit{Domain Blocks}. The original input image is partitioned into non-overlapping $R$ blocks of size $N$ $\times$ $N$ to form $R$ block set \{$R$\}.\\

\noindent \textit{Range Blocks}. The input image is also divided into overlapped $D$ blocks to form D block set  \{$D$\}, with size $2N$ $\times$ $2N$.\\

\noindent \textit{Transformation}. The contractive transformation $w$ is a combination of a geometrical transformation and luminance transformation. The transformed version of the domain block can be rotated, mirrored, contrast scaled or translated, so the transformation can be shown as an affine transformation.\\

\noindent All $D$ blocks are contract to the size $N$ $\times$ $N$ (the same size of $R$ blocks) by downsampling or averaging the intensities of its four neighboring pixels.

\noindent The main computational step in fractal image coding is the mapping of domains to range blocks; contracted $D$ block is extended through $8$ isometric transformations (including identical transformation; rotations by $\ang{90}$, $\ang{180}$ and $\ang{270}$  about the block center; symmetric reflections about central vertical axis, horizontal central axis, leading diagonal and secondary diagonal). The extended domain pool that constructs a codebook is defined as \{$\Tilde{D}$\}. In order to find the best matching block for each $R$ block, contraction mapping transformation $W_i$ is defined as:

\begin{equation}
W_i =
    \begin{bmatrix}
    x\\
    y\\
    z\\
\end{bmatrix}
+
    \begin{bmatrix}
    a_i&b_i&0\\
    c_i&d_i&0\\
    0&0&s_i\\
\end{bmatrix}
    \begin{bmatrix}
    x\\
    y\\
    z\\
\end{bmatrix}
+
    \begin{bmatrix}
    e_i\\
    f_i\\
    o_i\\
\end{bmatrix}
\end{equation}

\noindent where $x$ and $y$ represent spatial coordinates,  $z$ indicates pixel value; $a_i$, $b_i$, $c_i$ and $d_i$ define one of the $8$ isometric transformations; $s_i$ represents brightness adjustment factor with its absolute value smaller than $1$ and $o_i$ notes brightness offset factor.

\noindent This operation allows to find the best matching block from the extended domain pool \{$\Tilde{D}$\} for each $R$ block, satisfying the minimum distortion error $E(R,\Tilde{D})$ of Eq. (\ref{eq:2}).

\begin{equation} \label{eq:2}
  E(R,\Tilde{D}) = ||R-(s \cdot \Tilde{D} + o \cdot I)||
\end{equation}
    
\begin{figure}[!ht]
\centering
  \includegraphics[width=.5\textwidth]{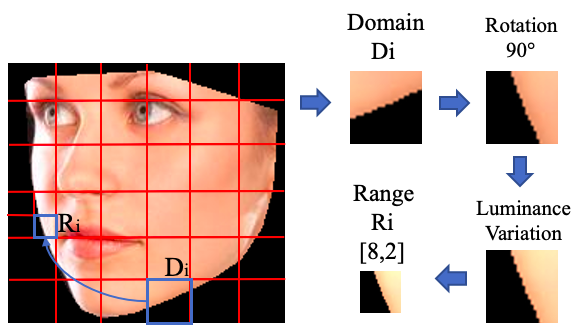}
  \caption{A detail of the fractal coding process, in particular the rotation and the lighting variation leading to the range located in row 8 and column 2.}
  \label{fig:codec_details}
\end{figure}

\noindent After fractal parameters of all $R$ blocks are stored as the result of compression, the total fractal encoding process is completed. In Figure~\ref{fig:codec1} we can see the transformation of the block of the domain in the block of the range. Images shown are from the dataset AFLW2000, described in Section V., and to which we have applied a mask, as described in Section~\ref{sec_met}. The two images are labeled with the same head pose, $\ang{5}$ Pitch, $\ang{30}$ Yaw and $\ang{-5}$ Roll. Due to the self-similarity induced by the fractal codec, we obtain that the same blocks of Domain will go in the same blocks of Range for both images within an acceptable margin of error Figure~\ref{fig:codec_details}. We use this propriety to build our method described in the following section.

\section{Method Description}
\label{sec_met}
\begin{figure*}
\centering
  \includegraphics[width=.9\textwidth]{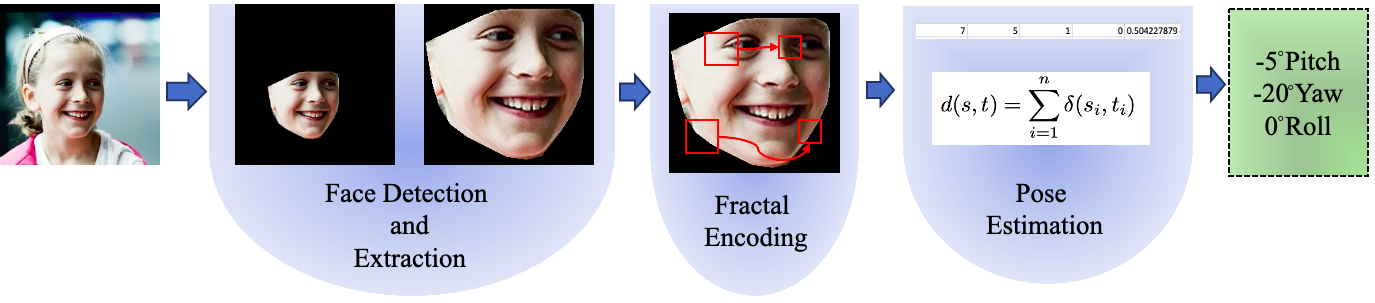}
  \caption{Workflow of the proposed method.}
  \label{fig:met}
\end{figure*}

HP\textsuperscript{2}IFS involves the use of fractal image coding algorithm to estimate face's pose and is structured in three main steps resumed below and depicted in Figure~\ref{fig:met}:
\begin{itemize}
    \item Step 1, \textit{Face detection and extraction};
\end{itemize}
\begin{itemize}
    \item Step 2, \textit{Fractal image coding algorithm} to obtain a matrix representing the fractal codes;
\end{itemize}
\begin{itemize}
    \item Step 3, \textit{Pose estimation}, transforming the fractal parameters in an array and comparing it to the angular array references obtaining in the same way (fractal codes) through Hamming distance metric.
\end{itemize}

More in detail, given an input image, a face localization algorithm is applied to detect key facial structures on the region of interest; a face alignment is made based on a group of detected facial landmarks, to realize the accurate detection of the face-ROIs. This step produce a masked image where non-face pixels are set to black.
Afterwards, the Viola-Jones algorithm \cite{violajones} is used to detect the presence and the location of the face in the image, resulting in a square region containing the face. The method resizes always the original image in 256 $\times$ 256 pixels. The fractal codes for all images are calculated using the original fractal image coding algorithm described in Section~\ref{sec_frac}. These fractal parameters contain information about luminance and geometrical position for the range blocks, the domain block corresponding to each range blocks and the geometrical transformation like rotation and resizing to match the domain block with the range block.
The algorithm transforms the parameters resulting from the fractal compression into an encoding-array, creating a model matrix. Finally, a comparison with the model is performed through the Hamming distance metric, which is particularly suited to compare two data strings having the same length and is defined as the number of bit positions in which the two bits are different according to:

\begin{equation} \label{eq:ham}
  d(s,t)=\sum_{i=1}^n \delta(s_i,t_i)
\end{equation}

where $s$ and $t$ are the strings to compare having lenght $n$ and $\delta(s_i,t_i)$ is the following function

\begin{equation} \label{eq:ham2}
  \delta(s_i,t_i)= 
  \begin{cases} 1, & \mbox{if }s_i\neq t_i \\ 
  0 & \mbox{if }s_i=t_i
\end{cases}
\end{equation}

This metric is notoriously simple to implement and fast to compute because its time complexity linearly depends on strings length. The minimum value obtained through the Hamming Distance between the model arrays and the input array give us the corresponding most similar head pose and its corresponding pitch, yaw and roll values.

\section{Experimental results}
As anticipated in the introduction, We relied on BIWI \cite{Biwi2011}  and AFLW2000 \cite{AFLW} datasets to perform the experiments. These two databases are completely different in terms of cooperativeness of the subjects and, as typically found in literature, we observed better results on BIWI in which subjects are cooperative. In order to test our method, as descibed in section~\ref{sec_met}, we manipulate the two datasets in two different ways to obtain the model set and the test set. \\
We tested HP\textsuperscript{2}IFS using a well known performance index for pose estimation, the Mean Absolute Error (MAE). MAE measures the average magnitude of the errors in a set of predictions and can be defined as follow:

\begin{equation} \label{eq:MAE}
  MAE=\frac{1}{n}\sum_{j=1}^{n} |y_j-\hat{y_j}|
\end{equation}
where $y_j$ is the ground truth, in our case the true angular value and $\hat{y_j}$ is the prediction, in our case the predicted angular value. We calculated the MAE for Pitch, Yaw and Roll separately and also an overall MAE of the error along the three axes.
In the following subsection we will describe results on BIWI and AFLW2000 in details.

\subsection{Results on BIWI}
\begin{figure}[!ht]
\centering
  \includegraphics[width=.45\textwidth]{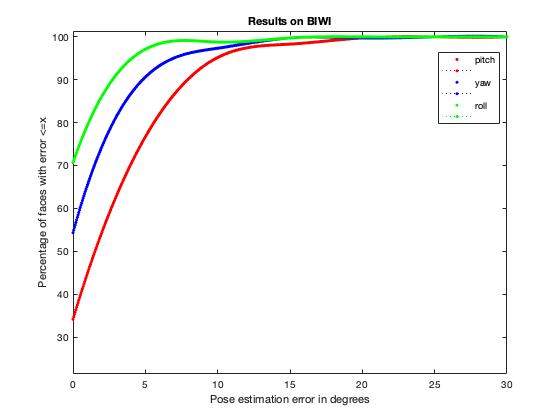}
  \caption{Errors on BIWI in terms of percentage of tested images.}
  \label{fig:ResBiwi}
\end{figure}

The dataset BIWI offers a lot of image per identity. For this reason we decided to create our model using a one-left-out technique. We created ten different set of experimental data using in turn one individual as tester and the others as model. In particular, we have to emphasise, that every subject has a wide range of poses that cover our angular variation in terms of pitch, yaw and roll. The results about the MAE for each subset tested are shown in table~\ref{table:BIWIsub}. As can be seen from the table, there are also great variations in errors between an experimental subset and the other. This can be explained by the geometrical difference between the faces of the subjects. As an example, there are few subjects with a long face and the majority of the subjects have a round face. When the former is the tester, the results appear worst than on a subjects with a round face because there are less similar sample of this shape of face in the model. This is confirmed by the use of Fractal Coding (See Section~\ref{sec_frac}) that emphasize self-similarity between the sectors of an image. 
The overall trend of the errors in terms of percentage of image tested can be seen in Figure~\ref{fig:ResBiwi}. We can observe a similar trend anticipated by the numerical results. In terms of Pitch, about 35\% of images have no error, 77\% of images have an error equal or less than $\ang{5}$, 95\% have an error equal or less than $\ang{10}$ and almost all images has an error less than $\ang{15}$. 
In terms of Yaw, results are more encouraging. About 55\% of images have no error, 90\% of images have an error equal or less than $\ang{5}$, 97\% have an error equal or less than $\ang{10}$ and almost all images has an error less than $1\ang{15}$. 
Roll achieved the best results because about 71\% of images have no error, 97\% of images have an error equal or less than $\ang{5}$, and almost all images have an error equal or less than $\ang{10}$. 

\begin{figure}[!ht]
\centering
  \includegraphics[width=.5\textwidth]{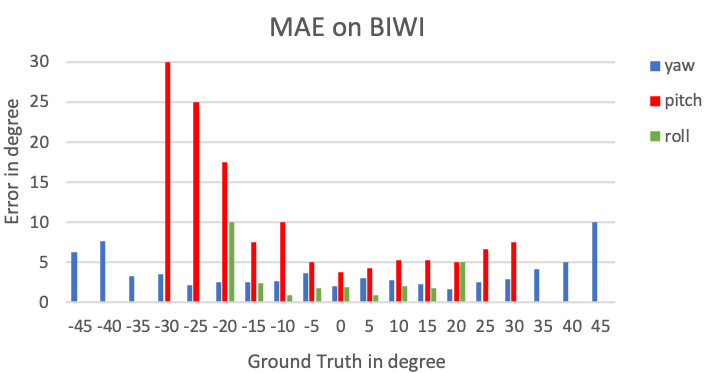}
  \caption{Errors on BIWI in terms of angular poses.}
  \label{fig:HistBiwi}
\end{figure}

Furthermore, we can analyse the errors in terms of angular variations. It is logical to assume that when head pose is more extreme, it is more difficult to estimate the angle accurately. From Figure~\ref{fig:HistBiwi} we can confirm that Pitch is the axis more affected by this issue, however the larger angular variation is in Yaw. In particular we can observe that, in general, the error decrease together with the estimated angle and increase when the estimated angle increase in modulus, confirming our hypothesis. In Table~\ref{tab:biwi} we show the comparison of HP\textsuperscript{2}IFS to state-of-the-art methods on BIWI. All methods consider are based on CNN or DL in general, with the exception of QT-PYR. Our method based on fractal coding achieves results comparable with recent techniques though they are based on machine learning approaches.

\begin{table}[h!]
\begin{center}
\caption{Results on the subsets of BIWI}
\label{table:BIWIsub}
\begin{tabular}{ |c|c|c|c|c| } 
\hline
Subset & Yaw & Pitch & Roll & overall MAE \\
\hline
1 & 5.85 & 10.97 & 3.55 & 6.79\\
2 & 4.27 & 5.84 & 2.27 & 4.12\\
3 & 4.13 & 5.45 & 11.84 & 7.14\\
4 & 2.83 & 5.1 & 1.68 & 3.2 \\
5 & 3.46 & 5.13 & 1.89 & 3.49\\
6 & 2.9 & 4.79 & 1.68 & 3.12\\
7 & 4.57 & 8 & 2.56 & 5.04\\
8 & 5.5 & 5.92 & 3 & 4.8\\
9 & 3.66 & 5.09 & 2.58 & 3.77 \\
10 & 3.33 & 6.02 & 1.96 & 3.77 \\
\textbf{mean} & \textbf{4.05} & \textbf{6.23} & \textbf{3.3} & \textbf{4.52} \\
\hline
\end{tabular}
\end{center}
\end{table}

\begin{table}[ht!]
\centering
\caption{Comparative results with the state-of-the-art on BIWI}
\label{tab:biwi}
\begin{tabular}{ |c|c|c|c|c| }
\hline
Method & Yaw & Pitch & Roll & MAE \\
\hline
Multi-Loss ResNet50~\cite{finegrained} & 5.17 & 6.97 & 3.39 & 5.177 \\ 
GPR~\cite{rasmussen2004gaussian} & 7.72 & 9.64 & 6.01 & 7.79 \\ 
PLS \cite{abdi2003partial}  & 7.35 & 7.87 & 6.11 & 7.11 \\ 
SVR \cite{smola2004tutorial} & 6.98 & 7.77 & 5.14 & 6.63 \\ 
hGLLiM \cite{drouard2017robust}  & 6.06 & 7.65 & 5.62 & 6.44 \\
QT PYR \cite{Ricciardi_pose} & 5.41 & 12.80 & 6.33 & 8.18 \\
FSA-Net \cite{yang2019fsa} & 4.27 & 4.96 & 2.76 & 3.996 \\
Coarse-to-Fine \cite{wang2019deep} & 4.76 & 5.48 & 4.29 & 4.84 \\ 
QuatNet \cite{hsu2018quatnet} & 4.01 & 5.49 & 2.93 & 4.14 \\
\hline
Proposed method & 4.05 & 6.23 & 3.30 & 4.52 \\
\hline

\end{tabular}
\end{table}

\subsection{Results on AFLW2000}
\begin{figure}[!ht]
\centering
  \includegraphics[width=.47\textwidth]{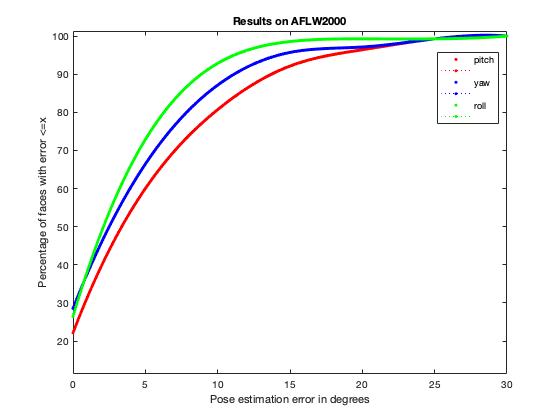}
  \caption{Errors on AFLW2000 in terms of percentage of tested images.}
  \label{fig:ResAFLW}
\end{figure}

The AFWL2000 dataset is considered very challenging to head pose estimation. In order to obtain a subdivision necessary for HP\textsuperscript{2}IFS, we used about 80\% of the dataset images randomly selected to build the model and the remaining to perform the tests. As observed also in the state-of-the-art, results on AFLW2000 are worse than results on BIWI. However, we can observe a behaviour of the error along the three axis similar to that achieved on BIWI. As can be seen in figure~\ref{fig:ResAFLW}, the errors on Roll are lower than errors on Pitch and Yaw, but they show a similar behaviour for small errors.

In particular, for both Pitch, Yaw and Roll, between 25\% and 30\% of images have no error. Then, the results began diversifying along the three axes. With regard to Pitch, 60\% of images have an error equal or less than $\ang{5}$, 81\% have an error equal or less than $\ang{10}$, 92\% have an error equal or less than $\ang{15}$, 96\% have an error equal or less than $\ang{20}$ and almost all images has an error less than $\ang{25}$. In terms of Yaw, 66\% of images have an error equal or less than $\ang{5}$, 87\% have an error equal or less than $\ang{10}$, 95\% have an error equal or less than $\ang{15}$, 97\% have an error equal or less than $\ang{20}$ and almost all images has an error less than $\ang{25}$. 
Finally, Roll, that shown also in this case the better results, 72\% of images have an error equal or less than $\ang{5}$, 92\% have an error equal or less than $\ang{10}$, 98\% have an error equal or less than $\ang{15}$, and almost all images has an error less than $\ang{20}$.

\begin{figure}[!ht]
\centering
  \includegraphics[width=.47\textwidth]{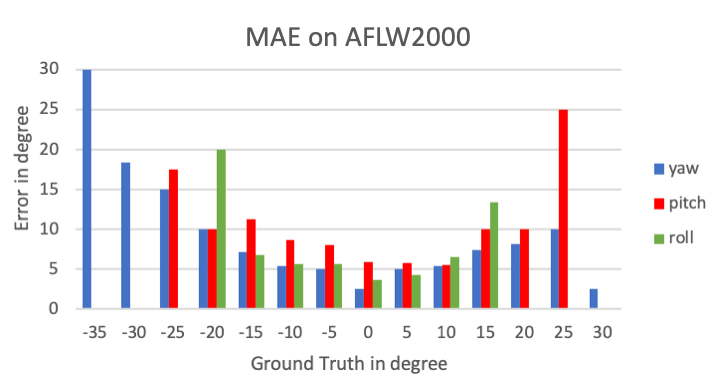}
  \caption{Errors on AFLW2000 in terms of angular poses.}
  \label{fig:HistAFLW}
\end{figure}

Also in this case, we can observe the behaviour of the error compared to the angular variations, in Figure~\ref{fig:HistAFLW}. The intervals of variation in degree for the ground truth depends on the available images in AFLW2000. Even if with different kind of errors, the average behaviour along the three axis is the same. The error increase together with the angle in modulus, just like we have observed in the results of BIWI. In Table~\ref{tab:aflw2000} the results of HP\textsuperscript{2}IFS on AFLW2000, compared to state-of-the-art methods, are shown. Even though the performance is lower then on BIWI, our results are comparable with other methods, encouraging us to consider the current implementation of fractal coding a promising starting point for further improvements of HP\textsuperscript{2}IFS also on competitive datasets as AFLW2000.

\begin{table}[ht!]
\centering
\caption{Comparative results with the state-of-the-art on AFLW2000}
\label{tab:aflw2000}
\begin{tabular}{|l|l|l|l|l|}
\hline
Method & Yaw & Pitch & Roll & MAE \\
\hline
Multi-Loss ResNet50~\cite{finegrained} & 6.470 & 6.559 & 5.436 & 6.155 \\ 
Hyperface \cite{ranjan2019hyperface} & 7.61 & 6.13 & 3.92 & 5.89 \\
KEPLER \cite{kumar2017kepler} & 6.45 & 5.85 & 8.75 & 7.01 \\
3DDFA \cite{zhu2017face}  & 5.400 & 8.530 & 8.250 & 7.393 \\ 
FAN \cite{bulat2017far} & 6.358 & 12.277 & 8.714 & 9.116 \\
QT PYR \cite{Ricciardi_pose} & 7.6 & 7.6 & 7.17 & 7.45 \\ 
QuatNet \cite{hsu2018quatnet}  & 3.973 & 5.615 & 3.92 & 4.503 \\ \hline
Proposed Method & 6.28 & 7.46 & 5.53 & 6.42 \\ \hline
\end{tabular}
\end{table}

\section{Conclusions}
We presented HP\textsuperscript{2}IFS, a novel method for head pose estimation based on partitioned iterated function systems to extract a fractal image code from input face to be matched against a reference array through Hamming distance. The proposed approach is explicitly is aimed at achieving accurate angular values for yaw, pitch and roll axis without requiring any neural network architecture and the related training overhead. According to the results of experiments carried out on both BIWI and AFLW2000 reference datasets, the reported pose estimation error is comparable to that of state-of-the-art methods based on machine learning and often inferior to that reported by the best performing non-training based methods.We are currently considering new non-linear transformations capable of enhancing the concept of self-similarity, as well as designing an ad-hoc metric for the distance between IFSs.

\bibliographystyle{plain}
\bibliography{Main}

\begin{thebibliography}{10}

\bibitem{Ricciardi_pose}
A.~F. {Abate}, P.~{Barra}, C.~{Bisogni}, M.~{Nappi}, and S.~{Ricciardi}.
\newblock Near real-time three axis head pose estimation without training.
\newblock {\em IEEE Access}, 7:64256--64265, 2019.

\bibitem{abdi2003partial}
Herv{\'e} Abdi.
\newblock Partial least square regression (pls regression).
\newblock {\em Encyclopedia for research methods for the social sciences},
  6(4):792--795, 2003.

\bibitem{barnsley1993fractal}
Michael Barnsley.
\newblock Fractal image compression.
\newblock {\em AK Peters}, 1993.

\bibitem{bulat2017far}
Adrian Bulat and Georgios Tzimiropoulos.
\newblock How far are we from solving the 2d \& 3d face alignment problem? (and
  a dataset of 230,000 3d facial landmarks).
\newblock In {\em The IEEE International Conference on Computer Vision (ICCV)},
  Oct 2017.

\bibitem{Czup2014}
B.~{Czupryński} and A.~{Strupczewski}.
\newblock High accuracy head pose tracking survey.
\newblock In {\em International Conference on Active Media Technology}, Aug
  2014.

\bibitem{distasi2005range}
Riccardo Distasi, Michele Nappi, and Daniel Riccio.
\newblock A range/domain approximation error-based approach for fractal image
  compression.
\newblock {\em IEEE Transactions on Image Processing}, 15(1):89--97, 2005.

\bibitem{drouard2017robust}
V.~{Drouard}, R.~{Horaud}, A.~{Deleforge}, S.~{Ba}, and G.~{Evangelidis}.
\newblock Robust head-pose estimation based on partially-latent mixture of
  linear regressions.
\newblock {\em IEEE Transactions on Image Processing}, 26(3):1428--1440, 2017.

\bibitem{Biwi2011}
G.~{Fanelli}, T.~{Weise}, J.~{Gall}, and L.~{Van Gool}.
\newblock Real time head pose estimation from consumer depth cameras.
\newblock In {\em Joint Pattern Recognition Symposium}, 2011.

\bibitem{fisher1995}
Yuval Fisher.
\newblock Fractal image compression: theory and application, 1995.

\bibitem{hsu2018quatnet}
Heng-Wei Hsu, Tung-Yu Wu, Sheng Wan, Wing~Hung Wong, and Chen-Yi Lee.
\newblock Quatnet: Quaternion-based head pose estimation with multiregression
  loss.
\newblock {\em IEEE Transactions on Multimedia}, 21(4):1035--1046, 2018.

\bibitem{kumar2017kepler}
A.~{Kumar}, A.~{Alavi}, and R.~{Chellappa}.
\newblock Kepler: Keypoint and pose estimation of unconstrained faces by
  learning efficient h-cnn regressors.
\newblock In {\em 2017 12th IEEE International Conference on Automatic Face
  Gesture Recognition (FG 2017)}, pages 258--265, May 2017.

\bibitem{trivedi2009}
E.~{Murphy-Chutorian} and M.~M. {Trivedi}.
\newblock Head pose estimation in computer vision: A survey.
\newblock {\em IEEE Transactions on Pattern Analysis and Machine Intelligence},
  31(4):607--626, Jan 2009.

\bibitem{ranjan2019hyperface}
R.~{Ranjan}, V.~M. {Patel}, and R.~{Chellappa}.
\newblock Hyperface: A deep multi-task learning framework for face detection,
  landmark localization, pose estimation, and gender recognition.
\newblock {\em IEEE Transactions on Pattern Analysis and Machine Intelligence},
  41(1):121--135, Jan 2019.

\bibitem{rasmussen2004gaussian}
Carl~Edward Rasmussen.
\newblock {\em Gaussian Processes in Machine Learning}, pages 63--71.
\newblock Springer Berlin Heidelberg, Berlin, Heidelberg, 2004.

\bibitem{finegrained}
Nataniel Ruiz, Eunji Chong, and James~M. Rehg.
\newblock Fine-grained head pose estimation without keypoints.
\newblock In {\em The IEEE Conference on Computer Vision and Pattern
  Recognition (CVPR) Workshops}, June 2018.

\bibitem{smola2004tutorial}
Alex~J. Smola and Bernhard Sch{\"o}lkopf.
\newblock A tutorial on support vector regression.
\newblock {\em Statistics and Computing}, 14(3):199--222, Aug 2004.

\bibitem{violajones}
Paul Viola and Michael Jones.
\newblock Rapid object detection using a boosted cascade of simple features.
\newblock In {\em IEEE conference on Computer Vision and Pattern Recognition.
  CVPR 2001}, volume~1, pages I--I. IEEE, 2001.

\bibitem{wang2019deep}
Yujia Wang, Wei Liang, Jianbing Shen, Yunde Jia, and Lap-Fai Yu.
\newblock A deep coarse-to-fine network for head pose estimation from synthetic
  data.
\newblock {\em Pattern Recognition}, 94:196--206, 2019.

\bibitem{yang2019fsa}
Tsun-Yi Yang, Yi-Ting Chen, Yen-Yu Lin, and Yung-Yu Chuang.
\newblock Fsa-net: Learning fine-grained structure aggregation for head pose
  estimation from a single image.
\newblock In {\em IEEE Conference on Computer Vision and Pattern Recognition},
  pages 1087--1096, 2019.

\bibitem{2015hybrid}
Shiping Zhu, Dongyu Zhao, and Fengchao Wang.
\newblock Hybrid prediction and fractal hyperspectral image compression.
\newblock {\em Mathematical Problems in Engineering}, 2015, 2015.

\bibitem{AFLW}
X.~{Zhu}, Z.~{Lei}, H.~{Shi}, X.~{Liu}, and S.~Z. {Li}.
\newblock Face alignment across large poses: A 3d solution.
\newblock In {\em IEEE Conference on Computer Vision and Pattern Recognition},
  2016.

\bibitem{zhu2017face}
X.~{Zhu}, X.~{Liu}, Z.~{Lei}, and S.~Z. {Li}.
\newblock Face alignment in full pose range: A 3d total solution.
\newblock {\em IEEE Transactions on Pattern Analysis and Machine Intelligence},
  41(1):78--92, Jan 2017.

\end{thebibliography}

\end{document}